\documentclass[conference]{IEEEtran}

\IEEEoverridecommandlockouts                              

\usepackage{graphicx,subfig}

\usepackage{hyperref}
\usepackage[sorting=none, style=ieee]{biblatex}
\usepackage{booktabs}  
\usepackage{threeparttable} 
\usepackage{makecell}
\usepackage{graphicx}
\usepackage{multicol}
\usepackage{multirow}
\usepackage{feyn}
\usepackage{amsmath,amssymb,amsfonts}
\usepackage{ulem}
\usepackage{caption}
\usepackage{fancyhdr} 

\usepackage{diagbox} 
\usepackage{makecell} 
\usepackage{booktabs} 
\usepackage{algorithm}
\usepackage{algpseudocode}
\usepackage[scaled]{helvet} 
\usepackage{tabularx}
\usepackage[table]{xcolor}
\usepackage{subcaption}
\usepackage{balance}
\usepackage{bbding}
\usepackage{color}
\usepackage{float}
\usepackage[justification=raggedright]{caption}
\usepackage[margin=1in]{geometry} 
\usepackage{ragged2e}
\usepackage{colortbl}
\usepackage{caption}

\usepackage{tabularx} 

\usepackage{hyperref}
\hypersetup{
	colorlinks=true,
	linkcolor=cyan,
	filecolor=blue,      
	urlcolor=black,
	citecolor=green,
}

\usepackage{biblatex}

\title{\LARGE \bf
Task-Oriented Pre-Training for Drivable Area Detection
}

\author{Fulong Ma, Guoyang Zhao, Weiqing Qi, Ming Liu, and Jun Ma    
\thanks{Fulong Ma, Guoyang Zhao, Weiqing Qi, and Ming Liu are with the Robotics and Autonomous Systems Thrust, The Hong Kong University of Science and Technology (Guangzhou), Guangzhou, China. (email: \{fmaaf,gzhao492,wqiad\}@connect.hkust-gz.edu.cn, eelium@hkust-gz.edu.cn.)}%
\thanks{Jun Ma is with the Robotics and Autonomous Systems Thrust, The Hong Kong University of Science and Technology (Guangzhou), Guangzhou, China, and also with the Division of Emerging Interdisciplinary Areas, The Hong Kong University of Science and Technology, Hong Kong SAR, China. (email: jun.ma@ust.hk).}%
}

\begin{document}

\maketitle

\thispagestyle{empty}
\pagestyle{empty}

\begin{abstract}

\footnotesize Pre-training techniques play a crucial role in deep learning, enhancing models' performance across a variety of tasks. 
By initially training on large datasets and subsequently fine-tuning on task-specific data, pre-training provides a solid foundation for models, improving generalization abilities and accelerating convergence rates. 
This approach has seen significant success in the fields of natural language processing and computer vision. 
However, traditional pre-training methods necessitate large datasets and substantial computational resources, and they can only learn shared features through prolonged training and struggle to capture deeper, task-specific features. 
In this paper, we propose a task-oriented pre-training method that begins with generating redundant segmentation proposals using the Segment Anything (SAM) model. We then introduce a Specific Category Enhancement Fine-tuning (SCEF) strategy for fine-tuning the Contrastive Language-Image Pre-training (CLIP) model to select proposals most closely related to the drivable area from those generated by SAM. 
This approach can generate a lot of coarse training data for pre-training models, which are further fine-tuned using manually annotated data, thereby improving model's performance. 
Comprehensive experiments conducted on the KITTI road dataset demonstrate that our task-oriented pre-training method achieves an all-around performance improvement compared to models without pre-training (as shown in Fig. \ref{spider}). Moreover, our pre-training method not only surpasses traditional pre-training approach but also achieves the best performance compared to state-of-the-art self-training methods. The open-source project
can be found at \href{https://sites.google.com/view/task-oriented-pre-training}{https://sites.google.com/view/task-oriented-pre-training}. 

\end{abstract}

\section{INTRODUCTION}

Drivable area detection stands as a pivotal component in the advancement of autonomous driving technologies, serving as a key component for ensuring vehicle safety and navigation efficiency. 
The importance of accurate drivable area detection is underscored by its direct impact on the decision-making processes of autonomous vehicles, influencing path planning, obstacle avoidance, and overall vehicle behavior in real-world driving scenarios \cite{li2013sensor}. 
In recent years, a lot of approaches have been explored, ranging from traditional image processing techniques to learning-based methods for enhanced precision and adaptability \cite{qiao2021drivable,fan2020sne, chen2019progressive,chang2203fast,min2022orfd,milli2023multi,li2021end}. 
These methodologies promote the technological evolution within the field and continuously improve the performance of drivable area detection tasks from the perspectives of data input modalities, model architectures, and the incorporation of auxiliary tasks. 
In this paper, we will explore from a novel pre-training perspective to further improve the performance of drivable area detection models.

\begin{figure}[t]
    \setlength{\abovecaptionskip}{0pt}
    \setlength{\belowcaptionskip}{0pt}
    \centering
    \includegraphics[width=1.0\linewidth]{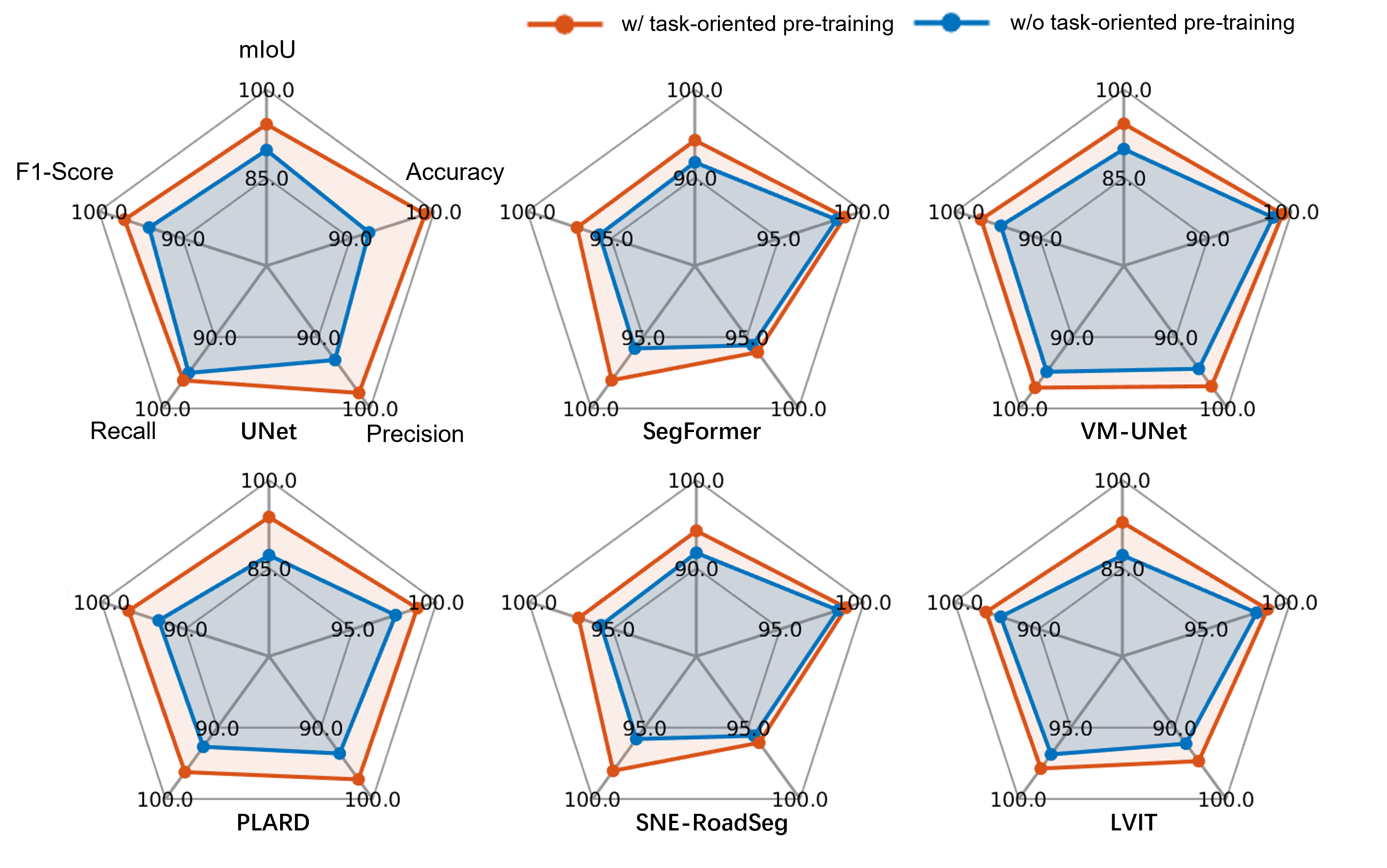}
    \captionsetup{font={footnotesize}}
    \caption{A visual qualitative comparison of the results using our task-oriented pre-training method versus without pre-training across several metrics: mIoU, Accuracy, Precision, Recall, and F1-Score.}
    \label{spider}
\end{figure}

\begin{figure*}[h]
    \setlength{\abovecaptionskip}{0pt}
    \setlength{\belowcaptionskip}{0pt}
    \centering
    \includegraphics[width=1.0\linewidth]{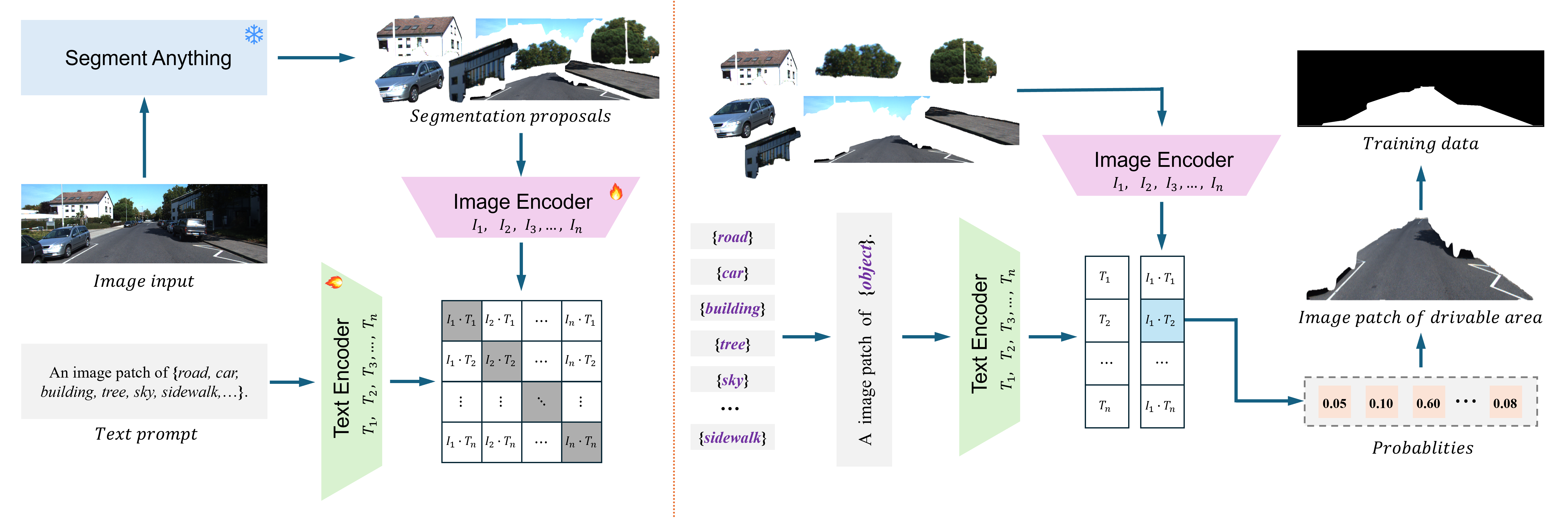}
    \captionsetup{font={small}}
    \caption{The overall architecture of our method.}
\label{arch}
\end{figure*}

Deep neural networks \cite{lecun1989backpropagation} have revolutionized the field of computer vision, pre-training techniques have undoubtedly made significant contributions\cite{ridnik2021imagenet,he2019rethinking}. 
In the pre-training and fine-tuning framework, models are first trained on a large-scale dataset (such as ImageNet \cite{russakovsky2015imagenet}) that may be unrelated to the target task, and then a secondary training are conducted on target tasks that often have limited amount of data. 
This pre-training and fine-tuning paradigm has led to advanced improvements in many computer vision tasks, including object image classification \cite{lu2007survey}, object detection \cite{zou2023object}, and semantic segmentation \cite{hao2020brief}. Similar to pre-training, self-supervised representation learning aims to learn general features on large-scale datasets and then fine-tune the model on target task. The difference lies in that self-supervised representation learning does not require any manually annotated data, instead, it relies on designing proxy tasks to complete the training. MoCo \cite{he2020momentum}, SimCLR \cite{chen2020simple}, MAE \cite{he2022masked}, and DINO \cite{caron2021emerging} are representative works.

Whether it is the pre-training or self-training techniques, both require substantial data and excellent computational resources to achieve desirable results. Moreover, both pre-training and self-training technologies are aimed at learning shared features, and it is challenging to learn deep features related to the target tasks.
Therefore, we are considering whether generating coarse training data for pre-training in the target data domain, followed by fine-tuning on training data with manually refined annotations, could achieve superior outcomes?
Fortunately, the development of foundational models in recent years has laid the solid foundation for validating our ideas.
In visual foundation models, SAM \cite{kirillov2023segment} and CLIP \cite{radford2021learning} represent two seminal works. SAM is a versatile segmentation model capable of generating precise masks for objects within a variety of images. CLIP performs image classification tasks by comprehending and correlating images with their language descriptions. Both models exhibit robust generalization capabilities.

In this paper, we integrate the SAM and CLIP models to propose a two-stage pre-training method for drivable area detection task. The overall framework is illustrated in Fig. \ref{arch}.
Briefly, the first stage generates redundant segmentation masks using SAM, and the second stage refines the CLIP model through an SCEF strategy to select the most appropriate masks from the segmentation masks.
In summary, our contributions are as follows:
\begin{itemize}

\item We propose a novel task-oriented pre-training framework for drivable area task, which achieves better performance improvements at an extremely low cost, as compared to traditional pre-training and self-training methods, it also offers insights for other tasks.

\item We propose a Specific Category Enhancement Fine-tuning (SCEF) strategy to fine-tune the CLIP model, enabling it to select the most appropriate mask from the redundant masks segmented by SAM.


\item We conduct comprehensive experiments on the KITTI road dataset, including experiments on models with different architectures and modalities. The experimental results demonstrate the effectiveness of our method.


\end{itemize}

\section{RELATED WORKS}

\subsection{Drivable Area Detection}
Drivable area detection is generally divided into image-based methods, point cloud-based methods, and multimodal methods. In image-based methods, they can be further divided into methods based on the front view and methods based on Bird's Eye View (BEV). In image-based methods, there are methods that detect obstacles in column pixels \cite{levi2015stixelnet} to obtain free space, as well as methods based on semantic segmentation \cite{liu2018segmentation}. In point cloud-based methods, they can be divided into traditional methods and deep learning-based methods. In traditional methods, the drivable area is usually determined based on the spatial structure information of the point cloud through geometric rules \cite{narksri2018slope,zermas2017fast,himmelsbach2010fast,lee2022patchwork++}. Learning-based methods include projecting point clouds onto a spherical surface, converting them into sphere images for use with 2D convolution methods \cite{wu2019squeezesegv2}, as well as methods that directly take point clouds as input for deep neural networks \cite{qi2017pointnet}. To fully utilize the information from multiple sensors, researchers have developed multimodal fusion methods \cite{fan2020sne,chen2019progressive,chang2203fast}, to improve algorithm performance. PLARD \cite{chen2019progressive} first converts point clouds into ADI images, then inputs the ADI images together with RGB images into a deep neural network for end-to-end learning. SNE-RoadSeg \cite{fan2020sne} integrates normal information and image information to detect drivable areas, while USNet \cite{chang2203fast} utilizes RGB images and binocular depth images combined with uncertainty estimation to achieve precise and efficient drivable area detection.

\subsection{Foundation Models}

Foundation models \cite{kirillov2023segment, liu2023grounding,} have revolutionized machine learning by providing generalized solutions capable of adapting to diverse tasks and domains with minimal fine-tuning \cite{awais2023foundational}. 
In the field of coputer vision, SAM \cite{kirillov2023segment} and CLIP \cite{radford2021learning} are two representative works. 
SAM is a versatile segmentation model that can generate accurate masks for objects in diverse images. 
CLIP aligns visual and textual representations through contrastive learning, enabling it to perform tasks like zero-shot image classification by understanding and relating images to natural language descriptions. 
Both models exemplify the integration of large-scale pre-training and generalization in artificial intelligence, contributing to the advancement of computer vision. 
In this work, we first use the SAM segmentation model to perform the ``everything” mode segmentation on the input image to obtain redundant segmentation proposals, 
then a specific category enhanced fine-tuning was used to select the segmentation result closest to the drivable area from the redundant segmentation proposals.
Finally, we collect a large amount of training data that combines SAM generation and CLIP selection for pre-training, and then fine-tune the model using manually annotated data to improve the performance of models for drivable area detection.

\subsection{Pre-Training and Self-Training}


Pre-training and fine-tuning strategies have been proven effective in a wide range of deep learning tasks.
A key driver of these effectiveness is the transfer learning paradigm \cite{pan2009survey}, where models are first pre-trained on large, diverse datasets to capture general features, and then fine-tuned on task-specific datasets to adapt these features to particular applications \cite{he2019rethinking,ridnik2021imagenet}. 
By leveraging the extensive knowledge acquired through large-scale pre-training, this process enhances performance on specific tasks, allowing models to not only generalize effectively but also to adeptly handle the subtleties of the target data.
Self-supervised representation learning does not require manual data labeling and learns features on large-scale datasets in a self-supervised manner. Then, by fine-tuning on downstream tasks, knowledge is transferred to these tasks.
MoCo \cite{he2020momentum}, SimCLR \cite{chen2020simple}, MAE \cite{he2022masked}, and DINO \cite{caron2021emerging} are representative works that have been proven effective across many computer vision tasks.

However, whether it is pre-training or self-supervised representation learning, both paradigms excel at transferring shared features to the target task \cite{zeiler2014visualizing}, but they fall short when it comes to capturing deeper, task-specific features that are more closely related to the objectives of the task.
Our proposed task-oriented pre-training method can focus on the target task itself, therefore, our method is capable of achieving better performance.

\section{METHOD}
\label{method}

It should be noted that the pre-training approach proposed in this paper differs from previous pre-training methods. Instead of pre-training on large unrelated datasets and then transferring knowledge to the target task through fine-tuning, we first pre-train the model on the target task by generating coarse training labels with the assistance of a SAM segmentation model. Subsequently, we fine-tune the model using precise manually annotated annotations from the target task. 
This approach enables the model to achieve superior performance, as the objectives of pre-training and fine-tuning are aligned within our pre-training framework.

Our method mainly comprises two steps. The first is to generate redundant mask proposals through SAM, and the second step involves selecting the most suitable one from these mask proposals through a fine-tuned CLIP model. As shown in Fig. \ref{arch}, on the left side of the dashed line, segmentation proposals are generated using the SAM model, which are then used to fine-tune the CLIP model. 
On the right side of the dashed line, the process involves the integration of segmentation proposals with text prompts to classify these image patches, thereby selecting the one that most closely approximates the drivable area.
Next, we will provide a detailed introduction to each module.
\begin{figure}[t]
    \setlength{\abovecaptionskip}{0pt}
    \setlength{\belowcaptionskip}{0pt}
    \centering
    \includegraphics[width=1.0\linewidth]{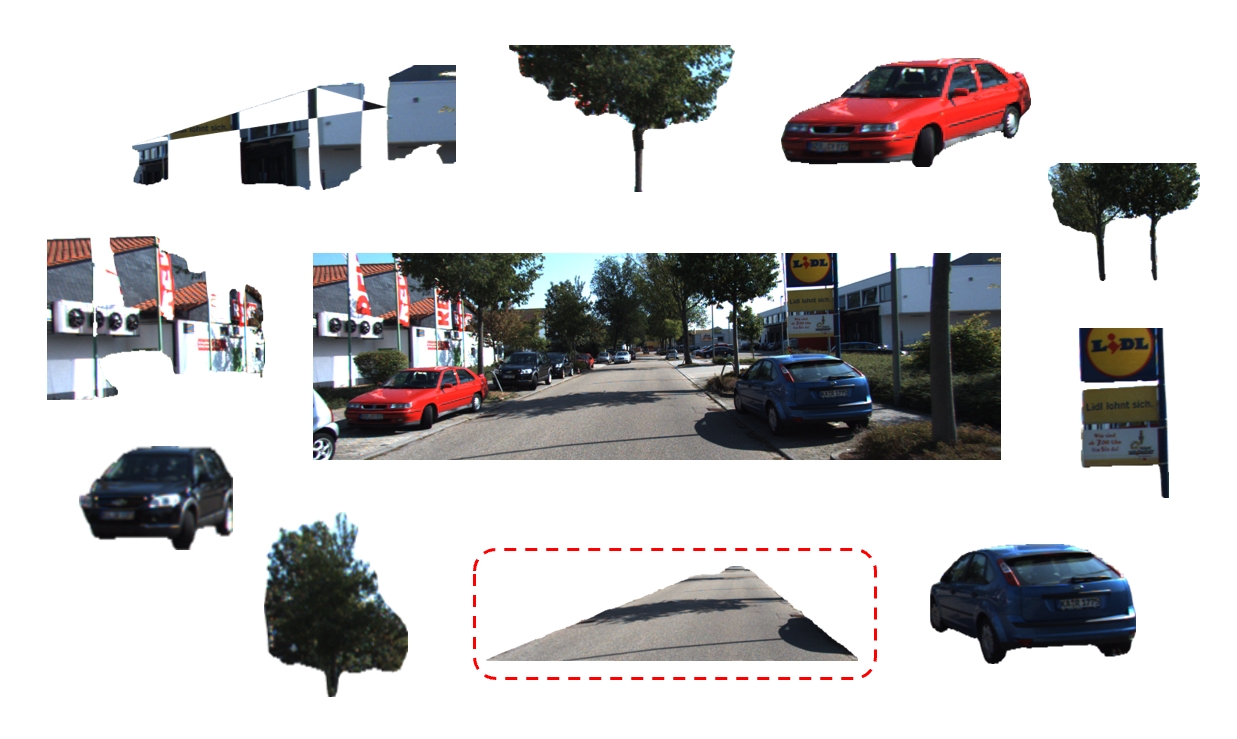}
    \captionsetup{font={footnotesize}}
    \caption{Segmentation results of the SAM model in the ``everything” mode, with the image patch within the red box representing pixels that belong to the drivable area.}
    \label{patches}
\end{figure}

\subsection{Redundant Masks Generation Using SAM}
The masks generated by SAM's ``everything" mode typically include segmentation mask for the object we want. As shown in Fig. \ref{patches}, the SAM model generates numerous image patches corresponding to different objects, and these image patches often contain what we most desire. For instance, the image patch within the red dashed box in Fig. \ref{patches} is the result we are looking for. In this step, we use the SAM model with frozen parameters to generate redundant mask proposals. 

\subsection{Specific Category Enhancement Fine-Tuning}

The KITTI road dataset officially annotates 289 images. We utilize these 289 annotated data and propose an SCEF strategy to fine-tune the CLIP model for more accurate selection of mask belonging to drivable area from redundant masks generated by SAM.
Specifically, SAM is first used to generate mask proposals on 289 annotated samples, and then we retain only the top 10 masks with the largest pixel area (as the drivable area tends to have a large pixel area).
Next, these 10 masks are input into a well-trained CLIP model to obtain the corresponding labels. Subsequently, the Intersection over Union (IoU) between these 10 masks and the drivable area ground truth manually annotated by the KITTI dataset is calculated. The label of the mask with the highest IoU is set as ``drivable area", while the labels of the remaining 9 masks remain unchanged as the zero-shot output from CLIP. The pseudo code for this process is shown in Algorithm \ref{alg1}.

\begin{figure*}[h]
    \setlength{\abovecaptionskip}{0pt}
    \setlength{\belowcaptionskip}{0pt}
    \centering
    \includegraphics[width=1.0\linewidth]{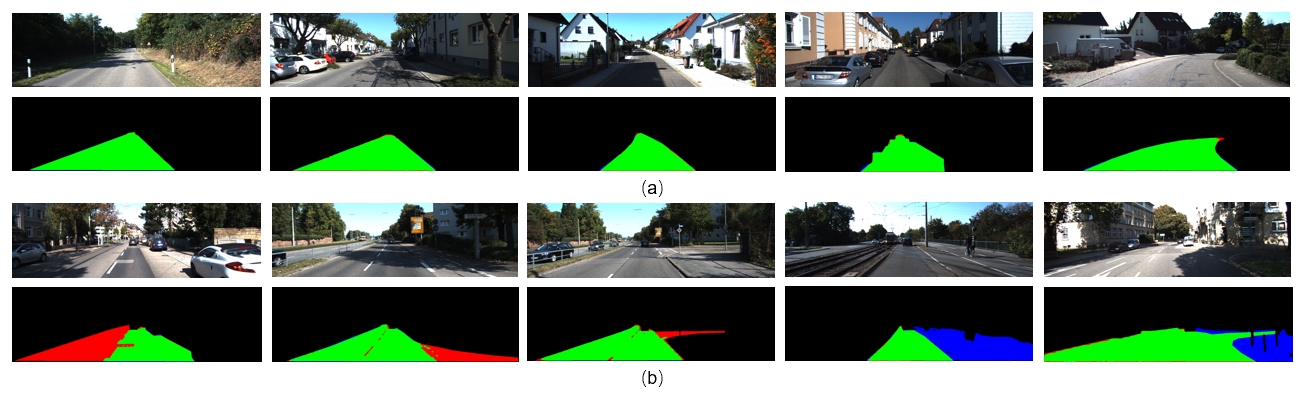}
    \captionsetup{font={scriptsize}}
    \caption{The qualitative visual comparison results between the training labels generated by our method and the KITTI road ground truth. Figure (a) shows some examples that are very close to the ground truth, while Figure (b) shows some failure cases.The true positive, false negative and false positive pixels are shown in green, red and blue, respectively.}
\label{vis_gt_label}
\end{figure*}

By using this strategy, we ensure that the class of the proposal generated by SAM that is closest to the drivable area in segmentation proposals is correct, while we do not focus on the classes of the remaining proposals. Subsequently, these segmentation proposals and their classes are used to fine-tune the CLIP model, thereby improving the accuracy of the fine-tuned CLIP model in selecting the mask belonging to the drivable area among those generated by SAM. 
The qualitative and quantitative results of the training data generated by our method are presented in Fig. \ref{vis_gt_label} and Table \ref{lable_vs_gt}, respectively.


\begin{algorithm}[t]
    \caption{Fine-Tuning Labels Generation} 
    \label{alg1}  
    \textbf{Input:}  Mask Proposals $\{M_{i}\}_{i=1}^N$, Mask Proposals' Classification Results $\{ C_{i} \}_{i=1}^{N}$, Annotation Mask of KITTI road Dataset $\hat{M}$.\\
    \textbf{Output:} Modified \{mask, category\} pairs $\{M_{i}^{\prime}, C_{i}^{\prime}\}_{i=1}^N $.
    \begin{algorithmic}[1]
    \State {\textbf{Initialization:}}
    \State $\{M\}_{i=1}^N:$ Get the segmentation masks of image $I$ using SAM model.
    \State $\{C\}_{i=1}^N:$ Get the classication results of $\{M\}_{i=1}^N$ using CLIP model.
    \State $\hat{M}:$ Read the annotation mask of image $I$ from disk.
    \State {\textbf{Main Loop:}}
    \For {$M_{i}$ in  $\{M\}_{i=1}^N$}
    \State Compute IoU score $S_{i}$ between $\hat{M}$ and $M_{i}$.
    \EndFor
    \State Replace the mask $ M_{i}^{\prime}$ in $\{M\}_{i=1}^N$ corresponding to the maximum value in $\{S_{i} \}_{i=1}^{N}$ with $\hat{M}$, and set the the corresponding class $ C_{i}^{\prime}$ in $\{ C_{i} \}_{i=1}^{N}$ to ``drivable area".
    \State Return the modified \{mask, category\} pairs $\{M_{i}^{\prime}, C_{i}^{\prime}\}_{i=1}^N $.
    \end{algorithmic}
\end{algorithm}

\begin{table}[t]
\renewcommand\arraystretch{1.2}
\captionsetup{font={footnotesize}}
\caption{The quantitative results of the accuracy, precision, recall, F1-score, and mIoU of the training labels generated by our method compared to the KITTI road ground truth.}
\label{lable_vs_gt}
\begin{center}
\setlength{\tabcolsep}{3.0mm}{
\begin{tabular}{c c c c c} 
\hline
 Accuracy &Precision &Recall & F1-score& mIoU \\
\hline
 98.34 &95.22& 95.43& 95.32 &91.07 \\
\hline
\end{tabular}
}
\end{center}
\end{table}

\section{EXPERIMENT}

\subsection{Experiment Details}



To ensure a comprehensive evaluation, we conducted extensive experiments across different neural network architectures and input modalities. Specifically, we selected three distinct neural network architectures—CNN, Transformer, and Mamba—corresponding to the algorithms UNet \cite{ronneberger2015u}, SegFormer \cite{fan2020sne}, and VM-UNet \cite{ruan2024vm}. Additionally, we chose three different data input modalities—RGB-LiDAR, RGB-Depth, and RGB-Language—corresponding to the algorithms PLARD \cite{chen2019progressive}, SNE-RoadSeg \cite{fan2020sne}, and LViT \cite{li2023lvit}. A total of six models were used to compare our pre-training method with the traditional pre-training approach on ImageNet. 
The LViT \cite{li2023lvit} model was originally designed for medical image segmentation tasks. However, to evaluate the effectiveness of our method with image-language modality input, we generated language branch inputs based on three different categories defined in the KITTI road dataset \cite{fritsch2013new}: ``urban unmarked", ``urban marked two-way road”, and ``urban marked multi-lane road”. 
These language descriptions, combined with the images, were then used to train and test the LViT model.

In addition, we conducted experiments to compare our pre-training method with four influential self-supervised representation learning methods: MoCo \cite{he2020momentum}, SimCLR \cite{chen2020simple}, MAE \cite{he2022masked}, and DINO \cite{caron2021emerging}. For these comparisons, we used ResNet-50 as the backbone network across all models, with UNet as the segmentation head.
It should be noted that, for the purpose of conducting comparative experiments, we obtained the trained ResNet-50 weight for MAE from reference \cite{tian2023designing}, in which, the ResNet-50 weight was trained using the self-supervised training paradigm like He \textit{et al.} proposed in the origin MAE \cite{he2022masked} paper.

\begin{table*}[h]
\renewcommand\arraystretch{1.0}
\captionsetup{font={footnotesize}}
\captionsetup{
  width=1.0\textwidth, 
}
\caption{The comparison results of three classic single-modal segmentation models on the KITTI road dataset for drivable area detection using our pre-training method, without any pre-training, and with conventional pre-training on ImageNet.}
\label{table_1}
\begin{center}
\setlength{\tabcolsep}{2.7mm}{
\begin{tabular}{c c c c c c c c} 
\toprule
Network  & Network Architecture    & Augmentation Method     & Accuracy &Precision &Recall & F1-score& mIoU \\

\hline
\multirow{3}{*}{U-Net \cite{ronneberger2015u}} &\multirow{3}{*}{CNN} 
& No Pre-training &97.22 &93.21 &95.03 &94.11 & 89.68 \\
 & &ImageNet Pre-training      &98.54  &96.58  &95.29  &95.93  &92.19  \\
 &  & Task-oriented pre-training     &\textbf{98.91} &\textbf{97.84} &\textbf{96.07} &\textbf{96.95} &\textbf{94.08} \\
\hline
\multirow{3}{*}{SegFormer \cite{xie2021segformer}} &\multirow{3}{*}{Transformer} 
& No Pre-training &98.46 &95.57 &95.80 &95.69 & 91.74 \\
 & &ImageNet Pre-training      &98.62  &\textbf{97.93}  &94.35  &96.11  &92.51  \\
 &  & Task-oriented pre-training & \textbf{98.93} &96.06 &\textbf{98.03} &\textbf{97.03} & \textbf{94.24} \\
\hline
\multirow{3}{*}{VM-UNet \cite{ruan2024vm}} &\multirow{3}{*}{Mamba} 

& No Pre-training & 97.77 & 94.45 & 94.87 & 94.66 & 89.86 \\
 & & ImageNet Pre-training  & 98.52 & 95.98 & 95.78 & 95.88 & 92.08 \\

 & & Task-oriented pre-training & \textbf{98.92} &\textbf{96.90} &\textbf{97.10} &\textbf{97.00} & \textbf{94.17} \\

\toprule
\label{single_modal}
\end{tabular}}
\end{center}
\end{table*}

\begin{table*}[h]
\renewcommand\arraystretch{1.0}
\captionsetup{font={footnotesize}}
\captionsetup{
  width=1.0\textwidth, 
}
\caption{
The comparison results of three multi-modal models with different inputs for drivable area detection using our pre-training method, without any pre-training, and with conventional pre-training on ImageNet.
}

\label{table_2}
\begin{center}
\setlength{\tabcolsep}{2.7mm}{
\begin{tabular}{c c c c c c c c} 
\toprule
Network  & Network Input    & Augmentation Method     & Accuracy &Precision &Recall & F1-score& mIoU \\

\hline
\multirow{3}{*}{PLARD \cite{chen2019progressive}} &\multirow{3}{*}{RGB + LiDAR} 
& No Pre-training &97.55 &93.63 &92.69 &93.16 & 87.20 \\
 & &ImageNet Pre-training      & 98.39 &95.73  & 95.29 &95.51  &91.41  \\
 &  & Task-oriented pre-training& \textbf{98.84} &\textbf{97.26} &\textbf{96.24} &\textbf{96.75} & \textbf{93.70} \\
\hline
\multirow{3}{*}{SNE-RoadSeg \cite{fan2020sne}} &\multirow{3}{*}{RGB + Depth} 
& No Pre-training &98.37 &95.64 &95.27 &95.46 & 91.31 \\
 & &ImageNet Pre-training      &98.85  &97.49  &96.09  &96.79  &93.78  \\
 &  & Task-oriented pre-training& \textbf{98.91} &\textbf{97.93} &\textbf{96.23} &\textbf{97.07} & \textbf{94.33} \\
\hline
\multirow{3}{*}{LViT \cite{li2023lvit}} &\multirow{3}{*}{RGB + Language} 
& No Pre-training &97.99 &92.25 &96.88 &94.51 &89.59  \\
 & &ImageNet Pre-training      &98.30  &93.68  & 97.06 &95.34  &91.10  \\
 &  & Task-oriented pre-training& \textbf{98.64} &\textbf{94.70} & \textbf{97.87} &\textbf{96.26} & \textbf{92.80} \\
\toprule
\label{multi_modal}
\end{tabular}}
\end{center}
\end{table*}







\subsection{Experiment Setup}

Our experiments are conducted in an Ubuntu 20.04 environment, equipped with an Intel i7 12700F CPU and a NVIDIA GeForce RTX 4090 GPU. 
For the generation of pre-training data, we selected data from the KITTI raw dataset that is 5 times the amount of the training set in the KITTI road dataset, totaling $289\times5=1445$ images.
Then, we use the method we propose to generate coarse training labels for these 1445 selected images, followed by pre-training with this data. For details on fine-tuning data, refer to Section \ref{dataset}.
We employed the PyTorch framework for model training, and for both pre-training and fine-tuning, the batch size is set to 2. The number of epochs for pre-training is 100, while for fine-tuning, it is 300.

\subsection{Dataset}
\label{dataset}
In our experiments, we use the publicly available KITTI road dataset \cite{fritsch2013new} to validate the effectiveness of our algorithm. The KITTI road dataset is one of the most popular and widely used datasets for road scene understanding and is commonly utilized for tasks such as road detection and lane line detection, it contains 289 frames training data and 290 frames testing data. When using the KITTI road dataset, it is often necessary to divide the dataset into three parts: training set, testing set, and validation set. However, the official KITTI road dataset only provides the training and testing portions and does not include a validation set. Therefore, researchers need to perform their own partitioning of the training and validation sets. Our partitioning method is as follows:
    \begin{itemize}
        \item  Training set, which consists of 173 images.
        \item  Validation set, which consists of 58 images.
        \item  Testing set, which consists of 58 images.
    \end{itemize}

\subsection{Evaluation Metrics}

Consistent with other drivable area detection works, we selected five commonly used evaluation metrics to assess the performance of our proposed method. These evaluation metrics are:
$Accuracy$, $Precision$, $Recall$, $F_{score}$ and $IoU$ (intersection over union),
and they were computed as follows:  $Accuracy =  \frac{N_{TP} + N_{TN}}{N_{TP} + N_{FP}  + N_{TN}  + N_{FN}}, Precision =  \frac{N_{TP}}{N_{TP}+N_{FP}}, Recall = \frac{N_{TP}}{N_{TP} + N_{FN}}, F1-score =  \frac{2 * Precision * Recall}{Precision + Recall}, IoU = \frac{N_{TP}}{ N_{TP} + N_{FP} + N_{FN}}$
where $N_{TP}$, $N_{TN} $, $N_{FP}$ and $N_{FN}$ represents the true positive, true negative, false positive, and false negative pixel numbers, respectively. 


\subsection{Performance Evaluation}

\subsubsection{Comparison with Pre-Training}
The comparison results between our method and the traditional ImageNet pre-training method are presented in Tables \ref{single_modal} and Table \ref{multi_modal}.
As shown in Table \ref{single_modal}, the three classic single-modality algorithms—UNet, SegFormer, and VM-UNet—using our pre-training method outperform the traditional ImageNet pre-training method on the KITTI road dataset. Specifically, compared to the traditional ImageNet pre-training method, UNet's F1-score improved from 95.93 to 96.95, an increase of 1.06\%, and its mIoU improved from 92.19 to 94.08, an increase of 2.05\%. For VM-UNet, the F1-score improved from 96.11 to 97.03, an increase of 0.96\%, and the mIoU improved from 92.51 to 94.24, an increase of 1.87\%. For SegFormer, the F1-score improved from 95.88 to 97.00, an increase of 1.17\%, and the mIoU improved from 92.08 to 94.17, an increase of 2.27\%. 
As shown in Table \ref{multi_modal}, the three multimodal algorithms with different input modalities—PLARD, SNE-RoadSeg, and LViT—also outperform the traditional ImageNet pre-training method on the KITTI road dataset when using our pre-training method. Specifically, for PLARD, the F1-score improved from 95.51 to 96.75, an increase of 1.3\%, and the mIoU improved from 91.41 to 93.70, an increase of 2.5\%. For SNE-RoadSeg, the F1-score improved from 96.79 to 97.07, an increase of 0.29\%, and the mIoU improved from 93.78 to 94.33, an increase of 0.58\%. For LViT, the F1-score improved from 95.34 to 96.26, an increase of 0.96\%, and the mIoU improved from 92.80 to 91.10, an increase of 1.85\%. Overall, our proposed task-oriented pre-training approach achieves promising performance improvements across both single-modality algorithms with different architectures and multimodal algorithms with various input modalities.

\subsubsection{Comparison with Self-Training}
The comparison experiments with four classic self-supervised representation learning methods—MoCo, SimCLR, MAE, and DINO—are presented in Table \ref{vs_self_training}. Compared to the four self-supervised methods, our proposed pre-training method overall outperforms these self-supervised representation learning methods. Specifically, the F1-score improved by 1.61\%, 1.32\%, 0.60\%, and 0.08\% compared to MoCo, SimCLR, MAE, and DINO, respectively. Similarly, the mIoU improved by 3.12\%, 2.54\%, 1.15\%, and 0.15\% compared to MoCo, SimCLR, MAE, and DINO, respectively.
Although our task-oriented pre-training method brings only a slight improvement compared to DINO, it is important to note that DINO requires training on two 8-GPU servers for approximately 3 days \cite{caron2021emerging}. In contrast, our task-oriented pre-training method requires only a single NVIDIA GeForce RTX 4090 GPU and takes less than 2 hours to train.
For specific tasks, our pre-training method is significantly more efficient and both resource- and energy-friendly.

\begin{table}[t]
\renewcommand\arraystretch{1.0}
\centering
\captionsetup{font={footnotesize}}
\captionsetup{
  width=0.52\textwidth, 
}
\tabcolsep=0.20cm
\fontsize{8}{10}\selectfont    
\caption{The comparison between our method and some self-training method like MoCo \cite{he2020momentum}, SimCLR \cite{chen2020simple}, MAE \cite{he2022masked}, and DINO \cite{caron2021emerging}. The best results are shown in bold type.}
\begin{tabular}{c|ccccc}
    \toprule
\diagbox [width=7em,trim=l] {Metrics}{Methods} & MoCo  & SimCLR & MAE &DINO & Ours  \\
\hline
Accuracy   &98.38&98.46& 98.71  & 98.88 & \textbf{98.91}  \\
Precision  &97.58&96.74& 97.40  & 97.47 & \textbf{97.84}  \\
Recall     &93.33&94.67& 95.37  & \textbf{96.28}& 96.07  \\
F1-Score   &95.41&95.69& 96.37  &  96.87& \textbf{96.95}  \\
mIoU       &91.23&91.75& 93.01  & 93.94 & \textbf{94.08}  \\
    \bottomrule
\end{tabular}\vspace{0cm}
    \label{vs_self_training}
\end{table}


\section{CONCLUSIONS}
\label{conclusions}

In this paper, we propose a task-oriented pre-training method that primarily consists of two steps. The first step involves processing the input images with a frozen-parameter SAM model to generate a lot of segmentation proposals for the objects in the image. The second step employs a CLIP model that has been fine-tuned using our proposed SCEF strategy to select the most appropriate mask from these proposals that belong to the drivable area, serving as coarse pre-training data.
We then use these generated coarse training data for pre-training models, followed by fine-tuning on the manually annotated KITTI road dataset. 
Our task-oriented pre-training method enables models to learn deeper and task-relevant features during the pre-training phase. In contrast, traditional pre-training and self-training methods are only able to learn some basic and shared features at the pre-training stage.
Finally, experiments on the KITTI road dataset demonstrate the effectiveness of this method, surpassing traditional pre-training on the ImageNet dataset as well as state-of-the-art self-training methods.
It is noteworthy that our method, in comparison to those pre-trained on ImageNet and self-training strategies, requires significantly lower amounts of data, computational resources, and training duration. This demonstrates that our approach is not only high-performing but also more efficient and cost-effective.

\normalem
\renewcommand*{\bibfont}{\footnotesize}
\printbibliography

@article{li2013sensor,
  title={A sensor-fusion drivable-region and lane-detection system for autonomous vehicle navigation in challenging road scenarios},
  author={Li, Qingquan and Chen, Long and Li, Ming and Shaw, Shih-Lung and N{\"u}chter, Andreas},
  journal={IEEE Transactions on Vehicular Technology},
  volume={63},
  number={2},
  pages={540--555},
  year={2013}
}

@article{li2021end,
  title={An end-to-end multi-task learning model for drivable road detection via edge refinement and geometric deformation},
  author={Li, Keqiang and Xiong, Hui and Yu, Dameng and Liu, Jinxin and Wang, Jianqiang and others},
  journal={IEEE Transactions on Intelligent Transportation Systems},
  volume={23},
  number={7},
  pages={8641--8651},
  year={2021}
}

@article{milli2023multi,
  title={Multi-modal multi-task (3mt) road segmentation},
  author={Milli, Erkan and Erkent, {\"O}zg{\"u}r and Y{\i}lmaz, As{\i}m Egemen},
  journal={IEEE Robotics and Automation Letters},
  year={2023}
}

@inproceedings{min2022orfd,
  title={{ORFD}: A dataset and benchmark for off-road freespace detection},
  author={Min, Chen and Jiang, Weizhong and Zhao, Dawei and Xu, Jiaolong and Xiao, Liang and Nie, Yiming and Dai, Bin},
  booktitle={IEEE International Conference on Robotics and Automation},
  pages={2532--2538},
  year={2022}
}

@INPROCEEDINGS{chang2203fast,
  author={Chang, Yicong and Xue, Feng and Sheng, Fei and Liang, Wenteng and Ming, Anlong},
  booktitle={IEEE International Conference on Robotics and Automation}, 
  title={Fast Road Segmentation via Uncertainty-aware Symmetric Network}, 
  year={2022},
  volume={},
  number={},
  pages={11124-11130}
}

@article{chen2019progressive,
  title={Progressive {LiDAR} adaptation for road detection},
  author={Chen, Zhe and Zhang, Jing and Tao, Dacheng},
  journal={IEEE/CAA Journal of Automatica Sinica},
  volume={6},
  number={3},
  pages={693--702},
  year={2019}
}

@inproceedings{qiao2021drivable,
  title={Drivable area detection using deep learning models for autonomous driving},
  author={Qiao, Donghao and Zulkernine, Farhana},
  booktitle={IEEE International Conference on Big Data},
  pages={5233--5238},
  year={2021}
}

@article{lecun1989backpropagation,
  title={Backpropagation applied to handwritten zip code recognition},
  author={LeCun, Yann and Boser, Bernhard and Denker, John S and Henderson, Donnie and Howard, Richard E and Hubbard, Wayne and Jackel, Lawrence D},
  journal={Neural Computation},
  volume={1},
  number={4},
  pages={541--551},
  year={1989},
  publisher={MIT Press}
}

@inproceedings{he2019rethinking,
  title={Rethinking {ImageNet} pre-training},
  author={He, Kaiming and Girshick, Ross and Doll{\'a}r, Piotr},
  booktitle={Proceedings of the IEEE/CVF International Conference on Computer Vision},
  pages={4918--4927},
  year={2019}
}

@article{ridnik2021imagenet,
  title={{ImageNet}-21k pretraining for the masses},
  author={Ridnik, Tal and Ben-Baruch, Emanuel and Noy, Asaf and Zelnik-Manor, Lihi},
  journal={arXiv preprint arXiv:2104.10972},
  year={2021}
}

@article{russakovsky2015imagenet,
  title={{ImageNet} large scale visual recognition challenge},
  author={Russakovsky, Olga and Deng, Jia and Su, Hao and Krause, Jonathan and Satheesh, Sanjeev and Ma, Sean and Huang, Zhiheng and Karpathy, Andrej and Khosla, Aditya and Bernstein, Michael and others},
  journal={International Journal of Computer Vision},
  volume={115},
  pages={211--252},
  year={2015},
  publisher={Springer}
}

@article{lu2007survey,
  title={A survey of image classification methods and techniques for improving classification performance},
  author={Lu, Dengsheng and Weng, Qihao},
  journal={International Journal of Remote Sensing},
  volume={28},
  number={5},
  pages={823--870},
  year={2007},
  publisher={Taylor \& Francis}
}

@article{zou2023object,
  title={Object detection in 20 years: A survey},
  author={Zou, Zhengxia and Chen, Keyan and Shi, Zhenwei and Guo, Yuhong and Ye, Jieping},
  journal={Proceedings of the IEEE},
  volume={111},
  number={3},
  pages={257--276},
  year={2023},
  publisher={IEEE}
}

@article{hao2020brief,
  title={A brief survey on semantic segmentation with deep learning},
  author={Hao, Shijie and Zhou, Yuan and Guo, Yanrong},
  journal={Neurocomputing},
  volume={406},
  pages={302--321},
  year={2020},
  publisher={Elsevier}
}

@inproceedings{kirillov2023segment,
  title={Segment anything},
  author={Kirillov, Alexander and Mintun, Eric and Ravi, Nikhila and Mao, Hanzi and Rolland, Chloe and Gustafson, Laura and Xiao, Tete and Whitehead, Spencer and Berg, Alexander C and Lo, Wan-Yen and others},
  booktitle={Proceedings of the IEEE/CVF International Conference on Computer Vision},
  pages={4015--4026},
  year={2023}
}

@inproceedings{levi2015stixelnet,
  title={{StixelNet}: A deep convolutional network for obstacle detection and road segmentation.},
  author={Levi, Dan and Garnett, Noa and Fetaya, Ethan and Herzlyia, Israel},
  booktitle={British Machine Vision Conference},
  volume={1},
  number={2},
  pages={4},
  year={2015}
}

@inproceedings{narksri2018slope,
  title={A slope-robust cascaded ground segmentation in {3D} point cloud for autonomous vehicles},
  author={Narksri, Patiphon and Takeuchi, Eijiro and Ninomiya, Yoshiki and Morales, Yoichi and Akai, Naoki and Kawaguchi, Nobuo},
  booktitle={International Conference on Intelligent Transportation Systems},
  pages={497--504},
  year={2018}
}

@inproceedings{zermas2017fast,
  title={Fast segmentation of {3D} point clouds: A paradigm on {LiDAR} data for autonomous vehicle applications},
  author={Zermas, Dimitris and Izzat, Izzat and Papanikolopoulos, Nikolaos},
  booktitle={IEEE International Conference on Robotics and Automation},
  pages={5067--5073},
  year={2017}
}

@inproceedings{himmelsbach2010fast,
  title={Fast segmentation of {3D} point clouds for ground vehicles},
  author={Himmelsbach, Michael and Hundelshausen, Felix V and Wuensche, H-J},
  booktitle={IEEE Intelligent Vehicles Symposium},
  pages={560--565},
  year={2010}
}

@inproceedings{lee2022patchwork++,
  title={Patchwork++: Fast and robust ground segmentation solving partial under-segmentation using {3D} point cloud},
  author={Lee, Seungjae and Lim, Hyungtae and Myung, Hyun},
  booktitle={IEEE/RSJ International Conference on Intelligent Robots and Systems},
  pages={13276--13283},
  year={2022}
}

@inproceedings{wu2019squeezesegv2,
  title={{SqueezeSegv2}: Improved model structure and unsupervised domain adaptation for road-object segmentation from a {LiDAR} point cloud},
  author={Wu, Bichen and Zhou, Xuanyu and Zhao, Sicheng and Yue, Xiangyu and Keutzer, Kurt},
  booktitle={International Conference on Robotics and Automation},
  pages={4376--4382},
  year={2019}
}

@inproceedings{fan2020sne,
  title={{SNE-RoadSeg}: Incorporating surface normal information into semantic segmentation for accurate freespace detection},
  author={Fan, Rui and Wang, Hengli and Cai, Peide and Liu, Ming},
  booktitle={European Conference on Computer Vision},
  pages={340--356},
  year={2020}
}

@article{liu2018segmentation,
  title={Segmentation of drivable road using deep fully convolutional residual network with pyramid pooling},
  author={Liu, Xiaolong and Deng, Zhidong},
  journal={Cognitive Computation},
  volume={10},
  pages={272--281},
  year={2018},
  publisher={Springer}
}

@inproceedings{qi2017pointnet,
  title={{PointNet}: Deep learning on point sets for {3D} classification and segmentation},
  author={Qi, Charles R and Su, Hao and Mo, Kaichun and Guibas, Leonidas J},
  booktitle={Proceedings of the IEEE/CVF Conference on Computer Vision and Pattern Recognition},
  pages={652--660},
  year={2017}
}

@inproceedings{radford2021learning,
  title={Learning transferable visual models from natural language supervision},
  author={Radford, Alec and Kim, Jong Wook and Hallacy, Chris and Ramesh, Aditya and Goh, Gabriel and Agarwal, Sandhini and Sastry, Girish and Askell, Amanda and Mishkin, Pamela and Clark, Jack and others},
  booktitle={International Conference on Machine Learning},
  pages={8748--8763},
  year={2021}
}

@inproceedings{caron2021emerging,
  title={Emerging properties in self-supervised vision transformers},
  author={Caron, Mathilde and Touvron, Hugo and Misra, Ishan and J{\'e}gou, Herv{\'e} and Mairal, Julien and Bojanowski, Piotr and Joulin, Armand},
  booktitle={Proceedings of the IEEE/CVF International Conference on Computer Vision},
  pages={9650--9660},
  year={2021}
}

@article{liu2023grounding,
  title={Grounding {DINO}: Marrying dino with grounded pre-training for open-set object detection},
  author={Liu, Shilong and Zeng, Zhaoyang and Ren, Tianhe and Li, Feng and Zhang, Hao and Yang, Jie and Li, Chunyuan and Yang, Jianwei and Su, Hang and Zhu, Jun and others},
  journal={arXiv preprint arXiv:2303.05499},
  year={2023}
}

@article{awais2023foundational,
  title={Foundational models defining a new era in vision: A survey and outlook},
  author={Awais, Muhammad and Naseer, Muzammal and Khan, Salman and Anwer, Rao Muhammad and Cholakkal, Hisham and Shah, Mubarak and Yang, Ming-Hsuan and Khan, Fahad Shahbaz},
  journal={arXiv preprint arXiv:2307.13721},
  year={2023}
}

@article{pan2009survey,
  title={A survey on transfer learning},
  author={Pan, Sinno Jialin and Yang, Qiang},
  journal={IEEE Transactions on Knowledge and Data Engineering},
  volume={22},
  number={10},
  pages={1345--1359},
  year={2009},
  publisher={IEEE}
}

@inproceedings{he2020momentum,
  title={Momentum contrast for unsupervised visual representation learning},
  author={He, Kaiming and Fan, Haoqi and Wu, Yuxin and Xie, Saining and Girshick, Ross},
  booktitle={Proceedings of the IEEE/CVF Conference on Computer Vision and Pattern Recognition},
  pages={9729--9738},
  year={2020}
}

@inproceedings{chen2020simple,
  title={A simple framework for contrastive learning of visual representations},
  author={Chen, Ting and Kornblith, Simon and Norouzi, Mohammad and Hinton, Geoffrey},
  booktitle={International Conference on Machine Learning},
  pages={1597--1607},
  year={2020}
}

@inproceedings{he2022masked,
  title={Masked autoencoders are scalable vision learners},
  author={He, Kaiming and Chen, Xinlei and Xie, Saining and Li, Yanghao and Doll{\'a}r, Piotr and Girshick, Ross},
  booktitle={Proceedings of the IEEE/CVF Conference on Computer Vision and Pattern Recognition},
  pages={16000--16009},
  year={2022}
}

@inproceedings{zeiler2014visualizing,
  title={Visualizing and understanding convolutional networks},
  author={Zeiler, Matthew D and Fergus, Rob},
  booktitle={European Conference on Computer Vision},
  pages={818--833},
  year={2014}
}

@article{li2023lvit,
  title={{LViT}: Language meets vision transformer in medical image segmentation},
  author={Li, Zihan and Li, Yunxiang and Li, Qingde and Wang, Puyang and Guo, Dazhou and Lu, Le and Jin, Dakai and Zhang, You and Hong, Qingqi},
  journal={IEEE Transactions on Medical Imaging},
  year={2023},
  publisher={IEEE}
}

@inproceedings{ronneberger2015u,
  title={{U-Net}: Convolutional networks for biomedical image segmentation},
  author={Ronneberger, Olaf and Fischer, Philipp and Brox, Thomas},
  booktitle={International Conference on Medical Image Computing and Computer-Assisted Intervention},
  pages={234--241},
  year={2015}
}

@article{ruan2024vm,
  title={{VM-UNet}: Vision {Mamba} {UNet} for medical image segmentation},
  author={Ruan, Jiacheng and Xiang, Suncheng},
  journal={arXiv preprint arXiv:2402.02491},
  year={2024}
}

@inproceedings{fritsch2013new,
  title={A new performance measure and evaluation benchmark for road detection algorithms},
  author={Fritsch, Jannik and Kuehnl, Tobias and Geiger, Andreas},
  booktitle={IEEE Conference on Intelligent Transportation Systems},
  pages={1693--1700},
  year={2013}
}

@article{tian2023designing,
  title={Designing bert for convolutional networks: Sparse and hierarchical masked modeling},
  author={Tian, Keyu and Jiang, Yi and Diao, Qishuai and Lin, Chen and Wang, Liwei and Yuan, Zehuan},
  journal={arXiv preprint arXiv:2301.03580},
  year={2023}
}

@article{xie2021segformer,
  title={{SegFormer}: Simple and efficient design for semantic segmentation with transformers},
  author={Xie, Enze and Wang, Wenhai and Yu, Zhiding and Anandkumar, Anima and Alvarez, Jose M and Luo, Ping},
  journal={Advances in Neural Information Processing Systems},
  volume={34},
  pages={12077--12090},
  year={2021}
}




\end{document}